\pdfoutput=1

\documentclass[11pt]{article}

\usepackage[final]{acl}

\usepackage{times}
\usepackage{latexsym}

\usepackage[T1]{fontenc}

\usepackage[utf8]{inputenc}

\usepackage{microtype}

\usepackage{inconsolata}

\usepackage{graphicx}

\usepackage{CJKutf8}
\usepackage{booktabs}
\usepackage{amsmath}
\usepackage{amssymb}

\title{Commonality and Individuality! Integrating Humor Commonality with Speaker Individuality for Humor Recognition}

\author{Haohao Zhu, Junyu Lu, Zeyuan Zeng, Zewen Bai, Xiaokun Zhang\thanks{Corresponding author}, 
\\{\bf Liang Yang, Hongfei Lin }\\
 School of Computer Science and Technology, Dalian University of Technology, China \\
\{zhuhh, dutljy, zengzeyuan, dlutbzw\}@mail.dlut.edu.cn \\
dawnkun1993@gmail.com, \{liang,hflin\}@dlut.edu.cn \\
}

\begin{document}
\begin{CJK}{UTF8}{gbsn} 

\maketitle
\begin{abstract}

Humor recognition aims to identify whether a specific speaker’s text is humorous. 
Current methods for humor recognition mainly suffer from two limitations: (1) they solely focus on one aspect of humor commonalities, ignoring the multifaceted nature of humor; and (2) they typically overlook the critical role of speaker individuality, which is essential for a comprehensive understanding of humor expressions. 
To bridge these gaps, we introduce the Commonality and Individuality Incorporated Network for Humor Recognition (CIHR), a novel model designed to enhance humor recognition by integrating multifaceted humor commonalities with the distinctive individuality of speakers.  
The CIHR features a Humor Commonality Analysis module that explores various perspectives of multifaceted humor commonality within user texts, and a Speaker Individuality Extraction module that captures both static and dynamic aspects of a speaker’s profile to accurately model their distinctive individuality. 
Additionally, Static and Dynamic Fusion modules are introduced to effectively incorporate the humor commonality with speaker’s individuality in the humor recognition process. 
Extensive experiments demonstrate the effectiveness of CIHR, underscoring the importance of concurrently addressing both multifaceted humor commonality and distinctive speaker individuality in humor recognition.

\end{abstract}

\section{Introduction}

Humor is a vital component of human communication and is crucial in the pursuit of creating human-like artificial intelligence.
As the basis for machines to understand and respond to humor, humor recognition has been a challenging problem for researchers because of its inherent subjectivity, fuzziness and semantic complexity \citep{annamoradnejad2024colbert} .
Despite these obstacles, the promise of developing human-like artificial intelligence has continually drawn researchers to this field \citep{weller2020rjokes, zeng2024leveraging} .

\definecolor{customgreen}{RGB}{151,208,119}
\definecolor{customyellow}{RGB}{255,153,51}
\begin{figure}[!t]
\centering
    \includegraphics[scale=0.65, trim=135 298 0 298, clip]{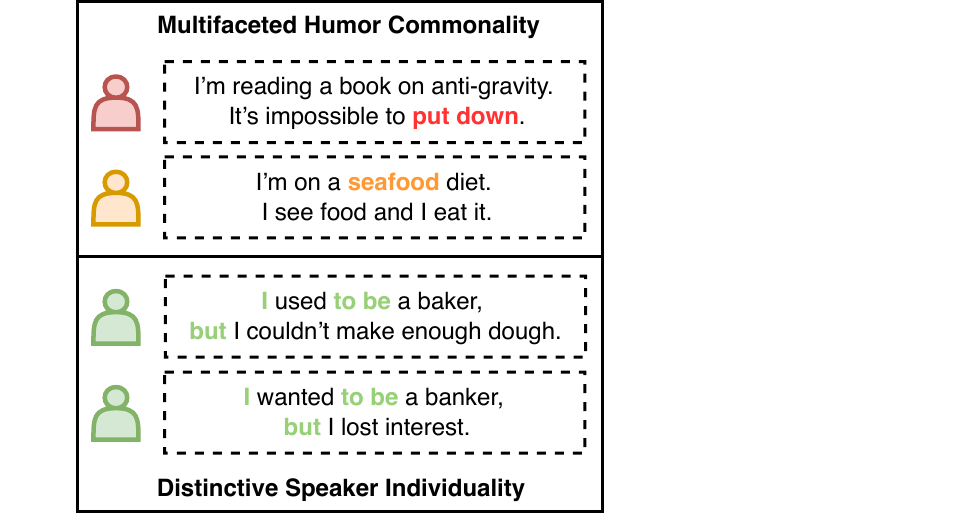}
\caption{Humor exhibits multifaceted commonalities, with \textcolor{red}{puns} highlighted in red and \textcolor{customyellow}{cognitive contradiction} in yellow. A speaker with distinctive individuality exhibits a \textcolor{customgreen}{unique humor expression style}, highlighted in green.}
\label{fig:intro}
\end{figure}

Current humor recognition methods employ a range of neural network architectures\citep{ren2022attention,annamoradnejad2024colbert, zeng2024leveraging} to automatically extract and learn semantic features for humor recognition.
Despite significant advancements in humor recognition, current methods still face two main issues.
Firstly, they typically focus on a single aspect of humor commonality in user text, neglecting the multifaceted nature of humor.
Humorous expressions often encompass various facets of commonality \citep{attardo2017humor}. 
For example, they typically include diverse elements like semantic puns, pragmatic implicit meanings, and cognitive contradictions, as illustrated in the upper part of Figure \ref{fig:intro}.
Overlooking these multifaceted characteristics of humor commonality can significantly limit a model’s ability to fully understand user-generated humor.
Secondly, existing methods often overlook the crucial role of speaker individuality.
Humor is not only dependent on language but is also deeply influenced by the speaker’s individuality such as social and cultural background, personality, and cognitive processes \citep{plessen2020humor, zhu2024integrating, zhu2024enhancing}. These characteristics shape each speaker’s unique humor expression style, as illustrated in the lower part of Figure \ref{fig:intro}. 
Neglecting these individualized characteristics can greatly diminish a model’s capability to effectively handle and interpret personalized expressions of humor.

Recognizing the need to address both the multifaceted commonality of humor expressions and the distinctive individuality of the speaker, we propose the Commonality and Individuality Incorporated Network for Humor Recognition (CIHR), 
which consists of four main components: Humor Commonalities Analysis, Speaker Individuality Extraction, Static Fusion and Dynamic Fusion.

Specifically, comprehensively analyze the commonality of humor from multiple perspectives, CIHR integrates a Humor Commonalities Analysis module that capitalizes on the interpretive capabilities of large language models (LLMs). 
This module employs prompts to direct LLMs in analyzing user language from six common perspectives of humor expression: semantic, pragmatic, syntactic, cultural, cognitive, and psychological.
Meanwhile, to accurately model a speaker’s distinctive individuality, the CIHR incorporates a Speaker Individuality Extraction module. 
This module captures both the static and dynamic information within a speaker’s profile, allowing for a more comprehensive and accurate modeling of individual traits. 
To effectively incorporate the commonalities of humor with the speaker’s individuality, CIHR is equipped with both Static and Dynamic Fusion modules. 
The Static Fusion module introduces Profile Guided Self-Attention (PG-SA) and Profile Guided Layer Normalization (PG-LN) to directly inject static profile information into the multi-perspective humor commonality analysis and dynamic profile text modeling processes. 
Meanwhile, the Dynamic Fusion module seamlessly integrates dynamic profile information into the multi-perspective humor commonality representation using attention mechanisms and gated neural networks.
Together, these modules enable CIHR to effectively combine the commonality of humor and the individuality of the speaker in the recognition process.

Extensive experiments were conducted, confirming the effectiveness of CIHR in the task of humor recognition. The results demonstrate the necessity and effectiveness of simultaneously considering both the commonalities of humor and the individual characteristics of speakers during the humor recognition process. The main contributions of this paper are summarized as follows:
\begin{enumerate}
	\item We deeply investigated the impact of both multifaceted humor commonality and distinctive speaker individuality on humor recognition, emphasizing the need for their integration to enhance this task.
	\item We introduced Commonality and Individuality Incorporated Network for Humor Recognition (CIHR), designed to effectively combine the commonality of humor and the individuality of the speaker in the recognition process.
	\item We conducted extensive experiments to validate the effectiveness of CIHR in humor recognition tasks, providing a detailed analysis of the impact and efficacy of each critical module within CIHR.
\end{enumerate}

\section{Related Work}

Early humor recognition research predominantly utilized feature engineering and traditional machine learning techniques to identify linguistic and stylistic features of humorous texts\cite{yang2015humor, liu2018modeling}.
For example, \citet{mihalcea2005making} employed features such as alliteration, antonyms, and adult slang to model humor, using classifiers like Naive Bayes (NB) and Support Vector Machines (SVM) \citep{cortes1995support} for humor recognition. \citet{raz2012automatic} explored grammatical, syntactic, semantic, and pragmatic features of humor texts, enriching the field of feature-based humor recognition. \citet{liu2018modeling} combined emotional and semantic analysis to model the emotional aspects of humor. 
While effective in some scenarios, these methods significantly struggle to capture the deep semantic and contextual nuances vital for understanding humor. 
Additionally, their reliance on manually designed features, which are labor-intensive to create and inflexible in application, further limits their effectiveness.

The advent of deep learning marked a significant shift towards using neural networks for automatic feature extraction. 
Additionally, the introduction of Transformer-based pretrained language models significantly enhanced the performance of humor recognition systems. 
For instance, \citet{chen2017predicting} developed a CNN-based model to recognize humor in speeches and puns, outperforming traditional methods even without manually designed features. 
\citet{weller2019humor} introduced the Transformer model, which further improved the ability to learn complex humor patterns and exhibited superior performance compared to other deep neural networks. 
\citet{xie2021uncertainty} utilized GPT \citep{radford2018improving} to calculate the inconsistency between punchlines in jokes, enhancing humor recognition based on these inconsistency scores. 
\citep{annamoradnejad2024colbert, yang2021choral} used pretrained BERT \citep{kenton2019bert} to obtain sentence embeddings for humor recognition. 
\citep{zeng2024leveraging} leveraged social network information and graph neural networks to enhance humor recognition results.

Despite significant progress in humor recognition research, recent approaches tend to focus primarily on single semantic layers, struggling to analyze user language from the diverse common perspectives of humor expression. Additionally, a critical limitation of these methods is their neglect of the speaker’s individual expression styles. 

\begin{figure*}[!ht]
\centering
    \includegraphics[scale=0.75, trim=30 222 0 287, clip]{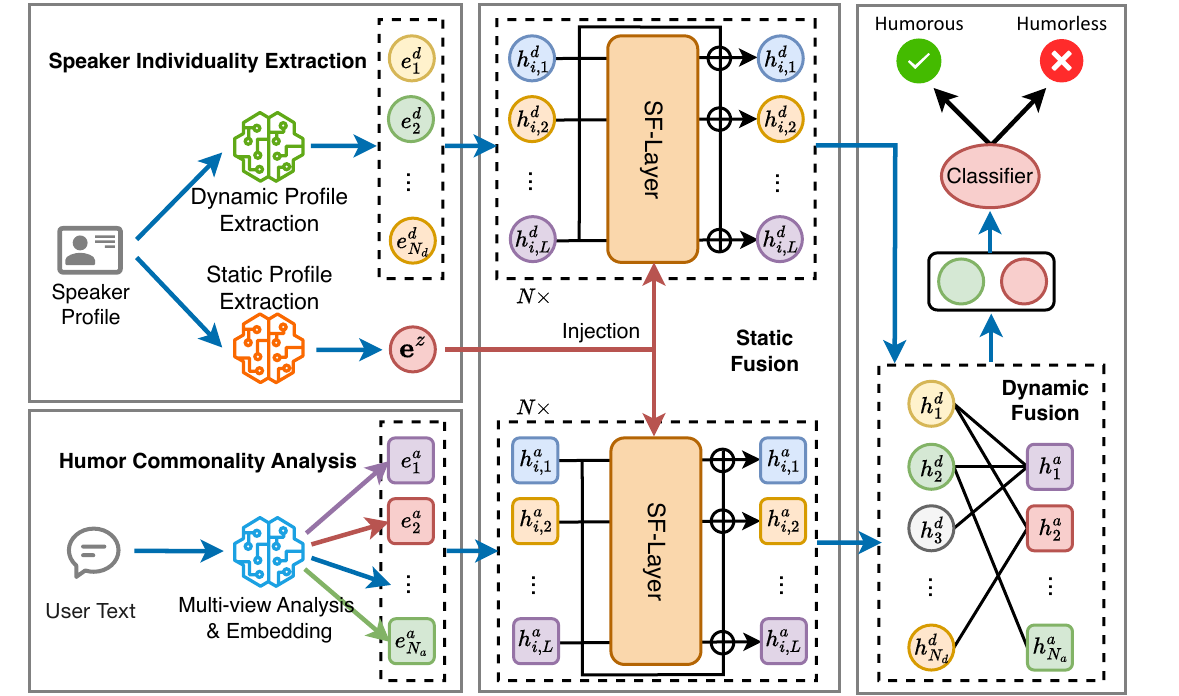}
\caption{The overall architecture of the proposed CIHR model, which consists of four main module: Humor Commonalities Analysis, Speaker Individuality Extraction, Static Fusion and Dynamic Fusion.}
\label{fig:model}
\end{figure*}

\section{Problem Statement}

The objective of humor recognition is to determine whether a specific speaker’s text $x$ is humorous or humorless. Formally, for a given speaker’s text $x$, the humor recognition model $y = f(x)$ predicts whether the text is humorous ($y=1$) or humorless ($y=0$).
As our approach aims to analyze user humor from both the perspectives of humor commonalities and speaker individuality, it is essential to incorporate additional user profile information. 
Accordingly, our inputs consist of the user text $x$ and the user profile $P_x$. Therefore, our humor recognition model is expressed as $y = f(x, P_x)$, where the user profile $P_x = \{S_x, D_x \}$ encompasses both static and dynamic aspects of the user profile.
The static profile $S_x$ contains stable attributes such as the user’s gender, age, region, and personal descriptions. Conversely, the dynamic profile $D_x$ consists of the user’s historical posts that are regularly updated.

\section{Methodology}

The overall architecture of CIHR is illustrated in Figure \ref{fig:model}. 
CIHR is structured around four main components: Humor Commonality Analysis module, Speaker Individuality Extraction module, Static Fusion module, and Dynamic Fusion module.

The Humor Commonality Analysis (HC) module leverages the powerful text interpretation capabilities of large language models to perform multi-perspective humor commonalities analyses for user texts. 
The Speaker Individuality Extraction (SI) module extracts both static and dynamic profiles from user profiles and embeds them to model the speaker’s individuality with greater detail and accuracy.
The Static Fusion (SF) module introduces Profile Guided Self-Attention (PG-SA) and Profile Guided Layer Normalization (PG-LN) to integrate information from the user’s static profile into the text modeling process. As a result, through the SF, CIHR obtains a Static Profile Guided multi-perspective humor text representation and a Static Profile Guided dynamic profile representation.
The \textbf{D}ynamic \textbf{F}usion (DF) Module leverages attention mechanisms and gated neural networks to produce a final text representation that effectively merges perspectives of commonality and individuality.

Finally, CIHR employs this dual-perspective text representation to determine whether user text is humorous. 
Detailed descriptions of each component follows in subsequent sections.

\subsection{Speaker Individuality Extraction}

The SI module consists of Static Profile Extraction (SP) and Dynamic Profile Extraction (DP) modules to extract and embed static and dynamic profiles from user profiles, respectively, enabling a comprehensive and precise modeling of user individuality.

\subsubsection{Static Profile Extraction}

The SP module processes textual and non-textual data separately within the static profile.
For textual information, represented as $S_x' = \{s_1, s_2, ... , s_n\}$, SP utilizes a standard text encoder BERT \citep{kenton2019bert} to derive its representation $Z_x' \in \mathbb{R}^{d'}$, where $d'$ denotes the feature dimension of textual information:
\begin{equation}
	Z_x' = \text{BERT}(S_x')
\label{eq:encoder}
\end{equation}

For non-textual information, represented as $S_x''$, SP normalizes it and employs a single-layer feed-forward neural network to achieve its embedded representation, yielding $Z_x'' \in \mathbb{R}^{d''}$, where $d''$ denotes the feature count of non-textual information:
\begin{equation}
	Z_x'' = \text{Norm}(S_x'') W_s + b_s,
\end{equation}
where $\text{Norm}(\cdot)$ denotes the normalization function, $W_s \in \mathbb{R}^{d'' \times d''}$, $b_s \in \mathbb{R}^{d''}$ are trainable parameters.

Finally, the textual features $Z_x'$ and non-textual features $Z_x''$ are concatenated and transformed into a unified static profile representation $\mathbf{e}^z \in \mathbb{R}^{d}$:
\begin{equation}
	\mathbf{e}^z = ( Z_x' || Z_x'') W_z + b_z,
\end{equation}
where $W_z \in \mathbb{R}^{(d'+d'') \times d}$ and $b_z \in \mathbb{R}^{d}$ are trainable parameters, and $(\cdot || \cdot)$ indicates concatenation along the final feature dimension.

\subsubsection{Dynamic Profile Extraction}

For each dynamic post in the user’s profile, denoted as $D_x^{(i)}$, we first tokenize the text into subword sequences $D_x^{(i)} = [d_1, d_2, ..., d_L ]$ using the WordPiece tokenizer \citep{kenton2019bert}, where $L$ is the maximum length of the text sequence. 
Special tokens [CLS] and [SEQ] are inserted at the beginning and end of each post, respectively. The token sequence is then converted into word embeddings $e^d_i = [e^d_{i,1}, e^d_{i,2}, ... , e^d_{i,L} ] \in \mathbb{R}^{L \times d}$ using embedding technology along with additional positional embeddings.

By aggregating these embeddings, we obtain the dynamic profile embedding representation $e^d = [e^d_1, e^d_2, ..., e^d_{N_d}] \in \mathbb{R}^{N_d \times L \times d}$, where $N_d$ represents the maximum number of historical texts in the dynamic profile.

\subsection{Humor Commonality Analysis}

To analyze humor expressions within user texts from various perspectives of humor commonalities, we harness the powerful interpretative abilities of LLMs (GPT-3.5). 
We have identified six humor commonality perspectives from psychological literature \citep{attardo2017humorandpra,attardo2017humor} and use prompts to guide the LLMs to interpret user texts from these distinct perspectives. The process of obtaining specific humor commonality perspectives through prompts is outlined in Equation \ref{eq:LLM}, with details on the six humor perspectives and corresponding prompts available in Appendix \ref{sec:a1}.
\begin{equation}
\label{eq:LLM}
	a_i' = \mathcal{L}(x, p_i),
\end{equation}
where $\mathcal{L}(\cdot)$ denotes the output of the LLM using prompts, and $p_i$ represents the prompt template for a specific perspective.
This process yields a set of humor analyses from $N_a$ perspectives, $A_x' = (a_1', a_2', ..., a_{N_a}')$, where  $N_a = 7$ indicates the number of identified humor commonality perspectives.
The CIHR subsequently concatenates each perspective’s analysis with the original text to create enhanced humor commonality texts $A_x = (a_1, a_2, ..., a_{N_a})$, where $a_i = (x || a_i')$.

Similar to the dynamic profile embedding process, 
we first tokenize the enhanced humor text $a_i$ into subword sequences $a_i = [a_{i,1}, a_{i,2}, \ldots, a_{i,L}]$ using the WordPiece tokenizer,
inserting [CLS] and [SEQ] tokens at the beginning and end of each text, respectively. 
Uniquely, an additional [SEQ] token is inserted between the original text $x$ and the analysis $a_i{\prime}$ to serve as a paragraph delimiter. 
In addition to word and positional embeddings, we incorporate segment embeddings \citep{kenton2019bert} to distinguish between the original text and the humor analysis. 
These embeddings are summed to convert the token sequence into word embeddings $e^a_i = [e^a_{i,1}, e^a_{i,2}, \ldots, e^a_{i,L} ] \in \mathbb{R}^{L \times d}$, 
thus forming a multi-perspective humor analysis embedding representation $e^a = [e^a_1, e^a_2, \ldots, e^a_{N_a}] \in \mathbb{R}^{N_a \times L \times d}$.
Overall, the process of obtaining multi-view humor commonality analysis embeddings is summarized in Equation \ref{eq:embedding2}:
\begin{equation}
\label{eq:embedding2}
\begin{aligned}
	e^a &= \text{Embedding}(A_x), \\
	A_x &= (a_1, a_2, ..., a_{N_a}), \\
	a_i &= (x || \mathcal{L}(x, p_i))
\end{aligned}
\end{equation}

\subsection{Static Fusion}

The Static Fusion Module is implemented by stacking multiple SF-Layers, 
designed to incorporate the user’s static profile information into the representations of both the user’s dynamic profile and the multi-perspective humor analysis.
This is achieved by guiding the entire text modeling process with the static profile.

\begin{figure}[!t]
\centering
    \includegraphics[scale=0.60, trim=160 270 0 270, clip]{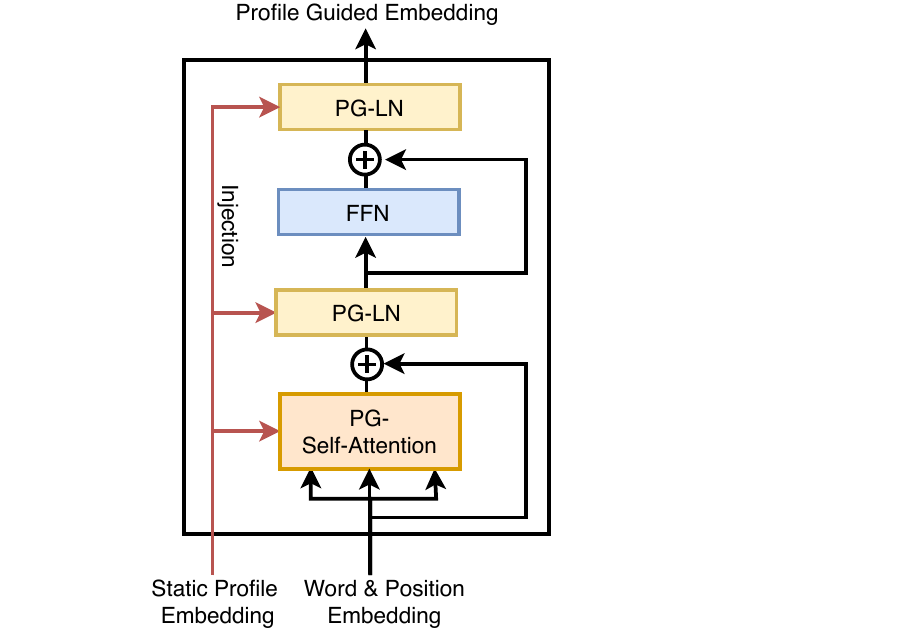}
\caption{Static Fusion Layer}
\label{fig:sfl}
\end{figure}

As illustrate in Figure \ref{fig:sfl}, the construction of SF-Layer is similar to a standard Transformer Encoder layer, which includes two main components: Multi-head Self-Attention and a Feed-Forward Network, both integrated through a Residual Connection and Layer Normalization for enhanced deep information modeling. 

The distinction lies in CIHR’s replacement of the standard Self-Attention and Layer Normalization with Profile Guided Self-Attention (PG-SA) and Profile Guided Layer Normalization (PG-LN).
This modification enables the model to incorporate static speaker profiles into the text modeling process.
By employing PG-SA and PG-LN, the SF Module effectively merges the user’s static profile with the text representation, resulting in a Static Profile-Guided humor commonalities analyses representations and Static Profile-Guided dynamic profile representations.

For a more detailed formulation of the SF-Layer, including PG-SA and PG-LN, please refer to Section \ref{sec:a3}.
The encoding process in the $n$-th Static Fusion Layer is depicted as follows:
\begin{equation}
\label{eq: layer}
\begin{aligned}
	\mathbf{h}^{(n)} &= \text{SF-Layer}^{(n)}
	(\mathbf{h}^{(n-1)}, \mathbf{e}^z),
\end{aligned}
\end{equation}
where $\mathbf{h}^{(n-1)} \in \mathbb{R}^{L \times d}$ is the output of the $n-1$th layer and the input to the $n$-th layer, 
and $h^{(0)}$ represents the initial embedding, either $e^d_i$ or $e^a_i$.
Overall, the text representation process utilizing n-layer Static Fusion Module based on PG-SA and PG-LN can be abstracted as shown in Equation \ref{eq:Static Fusion Module}:
\begin{equation}
\label{eq:Static Fusion Module}
\textbf{h}^{(n)} = \text{SF}(\textbf{h}^{(0)}, \mathbf{e}^z)
\end{equation}

Using this formula, we can derive representations for each user-specific post in the dynamic profile, $\mathbf{h}^d_{i} = \text{CLS}(\text{SF}(\mathbf{e}^d_{i}, \mathbf{e}^z))$, and for text combined with multi-perspective humor analysis, $\mathbf{h}^a_{i} = \text{CLS}(\text{SF}(\mathbf{e}^a_{i}, \mathbf{e}^z))$. 
Here, $\text{CLS}(\cdot)$ extracts the representation at the [CLS] token position, allowing us to collect complete user-specific dynamic profile features, $\mathbf{h}^d = [\mathbf{h}^d_{1}, \mathbf{h}^d_{2}, \ldots, \mathbf{h}^d_{N_d}]$, and multi-perspective text features combined with humor analysis, $\mathbf{h}^a = [\mathbf{h}^a_{1}, \mathbf{h}^a_{2}, \ldots, \mathbf{h}^a_{N_a}]$.

\subsection{Dynamic Fusion}

The Dynamic Fusion module is designed to further integrate the user’s dynamic profile into the text representation enriched with humor commonality analysis. 
Utilizing attention mechanisms and gated networks, it achieves a dual-perspective representation that merges both humor commonality and speaker individuality.

Drawing inspiration from \citep{lyu2023exploiting}, who enhanced sentiment analysis by incorporating user context through attention mechanisms, we first employ Multi-head Attention (MHA) \citep{vaswani2017attention} to derive text representations guided by the user’s dynamic profile. 
For the $\text{MHA}(Q, K, V)$, the multi-perspective humor analysis representation $\mathbf{h}^a$ serves as the query, and the dynamic profile representation $\mathbf{h}^d$ functions as both key and value:
\begin{equation}
\begin{aligned}
	\mathbf{h}^p &= \text{MHA}(\mathbf{h}^a, \mathbf{h}^d, \mathbf{h}^d) 
\end{aligned}
\end{equation}

To prevent the model from overly relying on dynamic profile information and neglecting current contextual data, 
a gated network is introduced to dynamically coordinate the influence of the dynamic profile guided humor analysis $\mathbf{h}^p$ and the multi-perspective humor analysis $\mathbf{h}^a$ on the final prediction. 
This gating mechanism allows for flexible learning from both representations to make the final humor assessment. 
Subsequently, we apply average pooling to aggregate the representations across multiple perspectives of humor analysis, yielding a final fused representation that embodies both humor commonality and speaker individuality:
\begin{equation}
\begin{aligned}
	\mathbf{h}^o &= \alpha \mathbf{h}^x + (1 - \alpha) \tilde{\mathbf{h}}^x, \\
	\alpha &= \sigma ([\mathbf{h}^x || \tilde{\mathbf{h}}^x] W + b), \\
\end{aligned}
\end{equation}
where $\mathbf{h}^x = \frac{1}{N_a} \sum_i^{N_a} \mathbf{h}_i^a, \tilde{\mathbf{h}}^x = \frac{1}{N_a} \sum_i^{N_a} \mathbf{h}_i^p$, with $W \in \mathbb{R}^{2d \times 1}$ and $b$ as a learnable parameters.

Through these mechanisms, the Dynamic Fusion module ensures that the final representation effectively integrates humor commonality and speaker-specific information, enhancing the model’s ability to accurately assess humor.

\subsection{Prediction and Objective}

Based on the final user text representation $\mathbf{h}^o$, which integrates both humor commonality and speaker individuality, we deploy a single-layer feed-forward network as a classifier to recognize humor:
\begin{equation}
	\hat{y}=\operatorname{softmax}\left(\mathbf{h}^o W_{o}+b_{o}\right),
\end{equation}
where $W_o \in \mathbb{R}^{d \times 2}$ and $b_o \in \mathbb{R}^{2}$ are trainable parameters.
 
During the training phase, we optimize the parameters of the CIHR by minimizing the cross-entropy loss, which quantifies the discrepancy between the predicted probability distribution and the actual humor labels. The objective function is defined as:
\begin{equation}
 	\mathcal{L}= \frac{1}{N_x} \sum_{i=1}^{N_x} ( y_{i} \log \left(\hat{y}_{i}\right)+\left(1-y_{i}\right) \log \left(1-\hat{y}_{i}\right) ),
\end{equation}
where $N_x$ represents number of training samples. 

\section{Experimental Settings}

\subsection{Dataset}

For effective humor recognition that integrates both humor commonalities and speaker individuality, our required dataset must include not only humorous texts with corresponding labels but also comprehensive user information, such as gender, age, region, and posting history. 
However, upon reviewing many popular humor recognition datasets, we found that they lack this essential user information. 
We discovered that the HumorWB dataset, mentioned in \citep{zeng2024leveraging}, includes not only labels identifying humor but also detailed user profiles, making it ideal for validating our approach.

Therefore, we selected the HumorWB dataset to evaluate our model and to test the feasibility and effectiveness of recognizing humor by considering both humor commonalities and speaker individuality.
The HumorWB dataset is sourced from Weibo, the largest Chinese social media platform, and includes 5,291 text entries from 2,151 users (of which 2,010 texts are marked as humorous) along with personal information such as the users’ gender, age, region, and historical posts. For detailed data statistics, please refer to \ref{sec:a4}.

\subsection{Baseline Methods}

To evaluate the performance of our CIHR model, we selected several advanced baseline models for comparison. These include both traditional machine learning models (SVM and LR) and deep learning models (TextCNN \citep{kim-2014-convolutional}, BiLSTM \citep{bertero2016long}, BERT \citep{kenton2019bert}, and the contemporary humor recognition models ANPLS \citep{ren2022attention}, ColBERT \citep{annamoradnejad2024colbert} and SCOG \citep{zeng2024leveraging}).
Additionally, we explored the performance of the latest large language models (Llama3-8b, Llama3.1-8b, OpenAI GPT-3.5, and GPT-4) on the task of humor recognition using prompts and included them as baseline methods to compare against our proposed CIHR model.
For details on the prompts used to guide the LLMs in humor recognition, please refer to \ref{sec:a2}.

\subsection{Implementation Details}

All hidden layer dimensions in the CIHR were set to 768. 
Within the Static Fusion Module, all parameters except those related to the user’s static profile were initialized using the pre-trained BERT model\footnote{\url{https://huggingface.co/hfl/chinese-bert-wwm-ext}}. 
All models were implemented using PyTorch and trained on a GeForce RTX 3090 GPU. 
The models were trained using the Adam optimizer \citep{kingma2015adam}, with a learning rate of 2e-5 for parameters initialized from the pre-trained BERT and 2e-3 for all other parameters. 
The batch size was set at 32, with training conducted for up to 20 epochs. 
The model that performed best on the validation set was selected for testing.

\begin{table}[!t]
\centering 
\caption{Overall results of CIHR and baseline models, with the best results in bold and the second-best results underlined. Statistical significance of pairwise differences for CIHR against the best baseline is determined by the t-test (p < 0.05).}
\begin{tabular}{lccccc}
\toprule[1.5pt]
\textbf{Approach} & \textbf{Acc} & \textbf{P} & \textbf{R} & \textbf{F1} \\
\midrule[1pt]
LR & 76.83 & 71.86 & 63.21 & 67.26 \\
SVM & 77.71 & 72.93 & 64.88 & 68.67 \\
TextCNN & 78.97 & 71.15 & 74.25 & 72.67 \\
BiLSTM & 76.45 & 65.73 & 78.26 & 71.45 \\
BERT & 75.19 & 79.96 & 71.98 & 75.76  \\
ANPLS & 78.72 & 79.33 & 74.45 & 76.81 \\
ColBERT & 82.87 & \underline{82.98} & 79.97 & \underline{81.45} \\
SCOG & \underline{83.00} & 76.11 & 79.93 & 77.98 \\
llama3-8b & 45.66 & 41.05 & \textbf{98.98} & 40.49 \\
llama3.1-8b & 65.59 & 53.75 & 76.74 & 65.45 \\
GPT-3.5 & 60.42 & 48.56 & \underline{91.24}& 63.39 \\
GPT-4 & 70.49 & 57.30 & 85.28 & 68.54 \\
\midrule[1pt]
CIHR(Ours) & \textbf{84.50} & \textbf{83.73} & 83.01 & \textbf{83.37} \\
\bottomrule[1.5pt]
\end{tabular}
\label{tb:overall results}
\end{table}

\section{Overall Performance}

As shown in Table \ref{tb:overall results}, CIHR outperforms all baseline models in accuracy and F1 score, highlighting its superior humor recognition capabilities. We believe the superior performance of CIHR stems from its ability to consider both the multifaceted commonalities of humor and the distinctive individuality of the speaker, which are crucial for comprehensively understanding humor.

Interestingly, we observed that all LLMs exhibit unusually high recall rates, particularly Llama3-8b and GPT-3.5. 
However, this result does not indicate superior humor recognition capabilities compared to CIHR, but rather a tendency to misclassify more humorless expressions as humorous. 
While GPT-4 shows improvement over GPT-3.5 and the Llama models, compared to other models, large language models still tend to overly interpret normal text as humorous. 
This lack of discernment accounts for their significantly lower performance in other metrics, particularly precision.

\begin{table}[!t]
\centering 
\caption{Results of various variants of the CIHR model.}
\begin{tabular}{lcccc}
\toprule[1.5pt]
\textbf{Approach} & \textbf{Acc} & \textbf{P} & \textbf{R} & \textbf{F1} \\
\midrule[1pt]
CIHR(Ours) & 84.50 & 83.73 & 83.01 & 83.37 \\
\midrule[1pt]
w/o HC & 83.50 & 82.54 & 82.13 & 82.33  \\
w/o SI & 83.25 & 82.16 & 83.59 & 82.87  \\
w/o SF & 83.75 & 83.02 & 83.40 & 83.21  \\
w/o DF & 83.88 & 82.72 & 83.69 & 83.20  \\
\bottomrule[1.5pt]
\end{tabular}
\label{tb:ablation}
\end{table}

\section{Detailed Analysis}

\subsection{Ablation Study}
\label{sec:ablation study}

The results presented above demonstrate the overall effectiveness of our CIHR model. 
To delve deeper into the impact of each critical module within CIHR, we conducted a series of ablation experiments, as shown in Table \ref{tb:ablation}. 
These experiments illustrate how each component contributes to the model’s performance.
For the variant labeled ‘w/o SF,’ we replaced the Static Fusion Module with the original pre-trained BERT. This substitution effectively removes the influence of the user’s static profile information, rather than eliminating the component entirely. In the ‘w/o DF’ configuration, we concatenated the user’s text representation with the dynamic profile and processed it through a single-layer feed-forward network, serving as a surrogate for the original Dynamic Fusion mechanism.

The results in Table \ref{tb:ablation} indicate that removing any part of the model leads to varying degrees of performance decline, highlighting the effectiveness and importance of each key module.
Specifically, removing the Humor Commonality Analysis module leads to a significant performance drop, underscoring the importance of analyzing text from humor commonality perspectives to understand textual humor effectively. 
The removal of the Speaker Individuality Extraction module also results in a notable decline in performance, suggesting that the individualized information of the speaker plays a significant role in analyzing the user’s humorous expressions.
The performance decreases observed upon removing the Static Fusion and Dynamic Fusion modules demonstrate that these modules effectively integrate personalized information into the text representation process, enhancing humor recognition capabilities. 

\subsection{Effect of Humor Commonality Analysis}

\begin{figure}[!t]
\centering
    \includegraphics[scale=0.23, trim=0 23 0 36, clip]{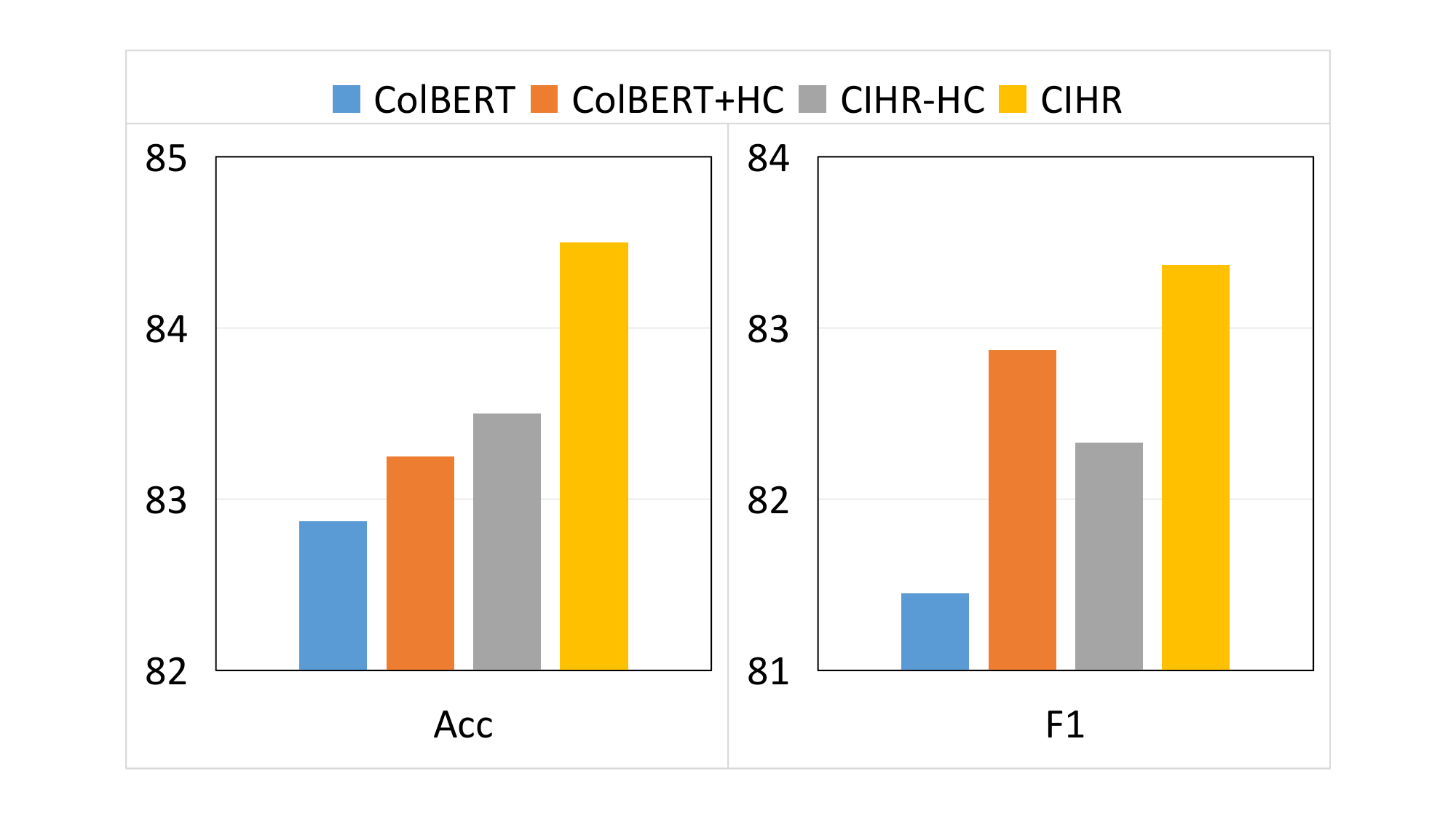}
\caption{Effect of Humor Commonality Analysis}
\label{fig:ablation1}
\end{figure}

To explore the impact of analyzing humor from various perspectives of its commonality, we compared the humor recognition performance of several model configurations: ColBERT, ColBERT enhanced with a Humor Commonality Analysis module (ColBERT+HC), CIHR minus the Humor Commonality module (CIHR-HC), and the full CIHR model. 
The results, illustrated in Figure \ref{fig:ablation1}, demonstrate that integrating the Humor Commonality Analysis module to the original ColBERT significantly improves performance. Conversely, removing this module from CIHR leads to a marked decline in effectiveness.

These findings robustly demonstrate the efficacy of the Humor Commonality Analysis module in enhancing humor recognition capabilities. More importantly, they confirm that analyzing humor from various perspectives of its commonality substantially boosts the model’s ability to comprehend and recognize humor, reinforcing the importance of this approach in the design of humor recognition systems.

\subsection{Effect of Speaker Individuality Extraction}

\begin{figure}[!t]
\centering
    \includegraphics[scale=0.23, trim=0 23 0 36, clip]{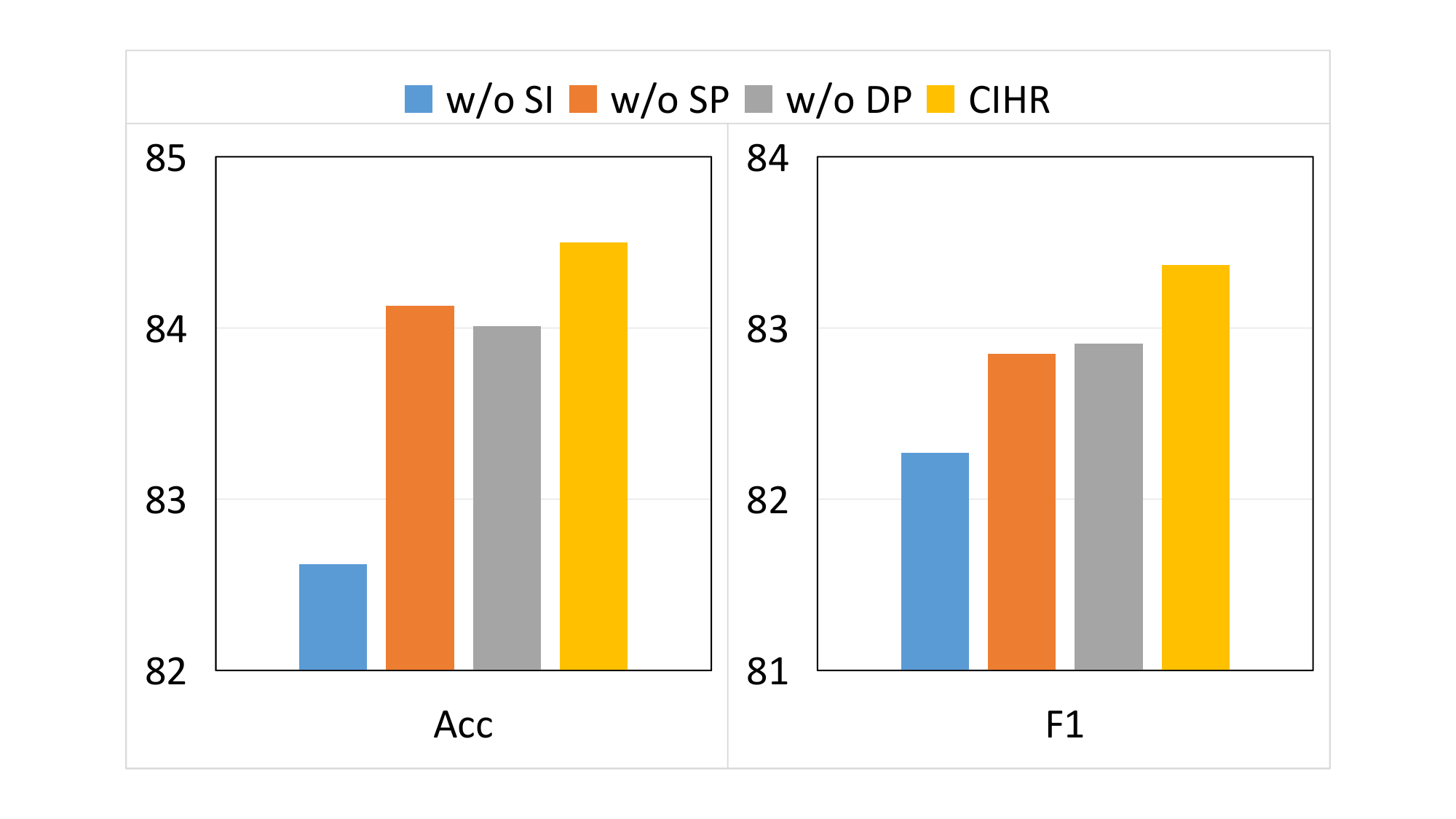}
\caption{Effect of Speaker Individuality Extraction}
\label{fig:ablation2}
\end{figure}

To explore the impact of leveraging user individuality on understanding and recognizing humor, 
we conducted experiments by selectively removing the Static Profile (SP) and Dynamic Profile (DP) from CIHR, 
as well as removing both simultaneously, and compared the humor recognition performance thereafter. 
The results, as depicted in Figure \ref{fig:ablation2}, show that ignoring either the Static Profile or the Dynamic Profile leads to a significant decrease in accuracy, recall, and F1 scores. 
Moreover, removing both profiles results in a dramatic decline in model performance.

These findings clearly demonstrate the critical role of user individuality information in enabling machines to understand and accurately recognize humor.
Ignoring these individual aspects restricts the machine’s ability to comprehend humor, thereby adversely affecting the overall effectiveness of humor recognition.

\subsection{Effect of Incorporating Both Humor Commonality and Speaker Individuality}

As illustrated in Figure \ref{fig:ablation1}, while considering either humor commonality alone (ColBERT+HC) or speaker individuality alone (CIHR-HC) does improve humor recognition performance compared to baseline method (ColBERT), integrating both dimensions elevates CIHR’s performance to a new level. 
This enhancement occurs because simultaneously addressing humor commonality and speaker individuality allows the machine to better understand humor expressions both from a general humor perspective and from the specific viewpoint of the speaker. Consequently, this dual approach significantly enhances humor recognition capabilities.
These results underscore the importance of considering both humor commonality and speaker individuality in the design of effective humor recognition systems, confirming that integrating these aspects leads to a more robust and accurate understanding of humor.

\section{Conclusion}

This study introduced the Commonalities and Individuality Incorporated Network for Humor Recognition (CIHR), an innovative approach addressing the limitations of existing models by incorporating humor commonalities with speaker individuality, aspects often neglected in traditional humor recognition.
CIHR features a Humor Commonality Analysis module to explore diverse aspects of multifaceted humor commonality within user texts, complemented by a Speaker Individuality Extraction module to capture unique individuality of each speaker.
The integration of these insights is enhanced through the Static and Dynamic Fusion modules, which effectively combine these elements to deepen the understanding of humor.
Experimental results validate CIHR’s effectiveness, showcasing significant improvements over existing models in terms of accuracy, precision, recall, and F1 scores. These findings emphasize the crucial importance of simultaneously considering both multifaceted humor commonality and distinctive speaker individuality to achieve robust and effective humor recognition.

\section*{Limitations}

Our approach uses user-specific information to enhance humor analysis by considering speaker individuality. 
Although validated on a specific dataset, 
the scarcity of datasets containing comprehensive user information limits its tested generalizability across different platforms and languages.
For a discussion on potential biases in the dataset we used, please refer to \ref{sec:a5}.
To address this, we are committed to expanding our research by collecting diverse datasets with user-specific information to thoroughly evaluate our method’s adaptability and effectiveness across various conditions.
Furthermore, our analysis of humor from various commonality perspectives is currently limited to using GPT-3.5, constrained by resource availability and the costs associated with API access. 
There is potential for improved analysis and enhanced humor recognition performance with the adoption of more advanced models, such as GPT-4. 

\section*{Ethics Statement}

The datasets utilized in our research were obtained from publicly accessible sources, the personal identifiers were totally anonymized, thereby safeguarding against the misuse of individual data.

\section*{Acknowledgments}

We thank reviewers for their comments, which provided some insights on this research that will further influence our future work. This research was supported by the National Natural Science Foundation of China (No. 62376051, 62076046).

\bibliography{PHD}

\clearpage

\appendix

\section{Appendix}
\label{sec:appendix}

\begin{CJK}{UTF8}{gbsn} 
\subsection{Perspectives of Humor Commonalities}
\label{sec:a1}

Since the beginning of the last century, there have been many work from different angles to analyze humor \citep{attardo2017humorandpra, attardo2017humor, attardo2017linguistics}, after collation and analysis, we summarized 6 different angles, hoping to guide large models through these theoretical support and empirical research, can more comprehensive and in-depth analysis of humor, improve model performance. These seven different angles can be divided into two aspects: language includes semantics, pragmatics and grammatical syntax, background includes culture, emotion, cognitive contradictions and psychology, the following are the six different angles and corresponding prompts we used in Speaker Individuality Extraction module:

Semantic analysis, Attardo's classic book Linguistic Theories of Humor details the semantic theory of humor, especially the polysemy of words and the role of puns in humor; The specific prompt is as follows,

\begin{quote}
\textit{“你是一位语言学专家，精通文本幽默分析。请分析以下文本中是否使用了词语多义性、双关语等修辞手法，并解释它们是否可能产生幽默效果。待分析文本: <用户文本>” \\
“You are a linguistics expert with expertise in textual humor analysis. Please analyze whether rhetorical devices such as polysemy and puns are used in the following text, and explain whether they may have a humorous effect. Text to be analyzed: <user text>”}
\end{quote}

Pragmatic analysis, while proposing the cooperation principal and the four maxims, Grice also pointed out that humor often arises from violations of these maxims, revealing how humor is achieved by cleverly manipulating language and context; The specific prompt is as follows,

\begin{quote}
\textit{“你是一位语言学专家，精通文本幽默分析。请根据上下文分析以下文本中是否存在隐含意义和言外之意，并判断它们是否构成了幽默。待分析文本: <用户文本>” \\
"You are a linguistics expert with expertise in textual humor analysis. Please analyze the following text according to the context to see if there are hidden meanings and subtexts, and determine whether they constitute humor. Text to be analyzed: {User text}”}
\end{quote}

Grammar and syntactic analysis, Veale from the perspective of natural language processing to explore the special use of syntactic structure in humor; The specific prompt is as follows,

\begin{quote}
\textit{"你是一位语言学专家，精通文本幽默分析。请检查以下文本的句子结构，分析是否存在不寻常的语法或句法现象，并判断它们是否为幽默服务。待分析文本: <用户文本>” \\
"You are a linguistics expert with expertise in textual humor analysis. Please check the sentence structure of the following text, analyze whether there are unusual grammatical or syntactic phenomena, and determine whether they serve humor. Text to be analyzed: <user text>”}	
\end{quote}

Cultural background analysis, Davies through the analysis of humor in different cultural backgrounds, explore how culture affects the understanding of humor; The specific prompt is as follows,

\begin{quote}
\textit{"你是一位语言学专家，精通文本幽默分析。请根据以下文本的文化或社会背景，分析是否有文化相关的幽默存在。待分析文本: <用户文本>” \\
"You are a linguistics expert with expertise in textual humor analysis. Please analyze whether there is culturally relevant humor based on the cultural or social background of the following text. Text to be analyzed: <user text>”}	
\end{quote}

Cognitive Contradiction Analysis, "Bisociative Thinking" by Koestler explains how cognitive contradictions play a role in humor; The specific prompt is as follows,

\begin{quote}
\textit{"你是一位语言学专家，精通文本幽默分析。请分析以下文本是否包含认知矛盾或不一致，并判断其是否用于制造幽默。待分析文本: <用户文本>” \\
"You are a linguistics expert with expertise in textual humor analysis. Please analyze whether the following text contains cognitive contradictions or inconsistencies, and determine whether it is used to create humor. Text to be analyzed: <user text>”}	
\end{quote}

Psychoanalysis, Freud explores humor and its subconscious foundations from a psychoanalytic perspective.
Through the above seven theoretical support and empirical research perspectives, We can study humor more scientifically and comprehensively. The specific prompt is as follows,

\begin{quote}
\textit{"你是一位语言学专家，精通文本幽默分析。请分析以下文本可能引发的心理反应，并判断它们是否能够产生幽默。待分析文本: <用户文本>” \\
"You are a linguistics expert with expertise in textual humor analysis. Please analyze the psychological reactions that the following texts may trigger and determine whether they can generate humor. Text to be analyzed: <user text>”}	
\end{quote}

\subsection{Prompts for LLMs Baselines}
\label{sec:a2}

The prompt guiding LLMs to determine whether the user text is humorous is as follows:

\begin{quote}
\textit{"你是一位语言学专家，精通文本幽默分析。请分析以下文本中是否会产生幽默效果，仅回复是/否。待分析文本: <用户文本>” \\
"You are a linguistics expert with expertise in textual humor analysis. Please analyze the following text to see if it creates humor, and respond yes/no only. Text to be analyzed: <user text>”}	
\end{quote}
\end{CJK}

\subsection{Details of Static Fusion Layer}
\label{sec:a3}

\subsubsection{Profile Guided Self-Attention}

Previous research indicates that adjustments in attention distribution can align sequence representations with attribute information \citep{zhang2021ma}, prompting the Static Fusion Module to inject user static profile information into the attention mappings.
Specifically, PG-SA first generates user-specific query vectors, leading to user-specific attention distributions. The $i$-th attention head for the PG-SA can be represented as:
\begin{equation}
\label{eq:head}
\begin{aligned}
\mathbf{h}_{(i)}' & =\operatorname{PG-SA}_{(i)}\left(\mathbf{h}, \mathbf{e}^z\right) \\
& =\operatorname{Softmax}\left(\frac{Q_{(i)}^{P} K_{(i)}^{\top}}{\sqrt{d}}\right) V_{(i)},
\end{aligned}
\end{equation}
where $Q_{(i)}^{P} \in \mathbb{R}^{N \times (d / K)}$ is the user-specific query matrix, and $K_{(i)}$, $V_{(i)} \in \mathbb{R}^{N \times (d / K)}$ are the key and value matrices, computed as follows:
\begin{equation}
\label{eq:qkv}
\begin{aligned}
	Q_{(i)}^{P} & =\operatorname{Linear}_{Q,(i)}(\mathbf{h}) \odot \sigma_{(i)}+\mu_{(i)} \\
	V_{(i)} & =\operatorname{Linear}_{V,(i)}(\mathbf{h}) \\
	K_{(i)} & =\operatorname{Linear}_{K,(i)}(\mathbf{h})
\end{aligned}
\end{equation}
where $\sigma{(i)}$ and $\mu_{(i)}$ are the scaling and shifting factors that modify the query matrix:
\begin{equation}
	\begin{aligned}
\sigma_{(i)} & =\mathbf{1}+\operatorname{Linear}_{\sigma,(i)}\left(\mathbf{e}^z\right) \\
\mu_{(i)} & =\operatorname{Linear}_{\mu,(i)}\left(\mathbf{e}^z\right)
\end{aligned}
\end{equation}

The final output of PG-SA, $\mathbf{h}{\prime} \in \mathbb{R}^{N \times d}$, is a concatenation of the outputs from all K heads:
\begin{equation}
	\mathbf{h}'=\operatorname{Linear}\left(\mathbf{h}_{(1)}' || \mathbf{h}_{(2)}' || \ldots || \mathbf{h}_{(K)}'\right)
\end{equation}
Notably, the user static profile embeddings $\mathbf{e}^z$ are zero-initialized (by zero-initializing $W_z$ and $b_z$) to ensure stability during model fine-tuning and to minimize the impact of unstable initial user profile embeddings on the learning of the model’s attention distribution.

\subsubsection{Profile Guided Layer Normalization}

Traditional layer normalization adjusts representations by learning shared gain and bias parameters across batches and sequences, treating them as contextualized representations. 
To adapt layer normalization to user-specific information, we introduce Profile Guided Layer Normalization (PG-LN), which is tailored to learn user-specific gain and bias factors. 
PG-LN is defined as follows:
\begin{equation}
\begin{aligned}
h_{i}' & =\left(\gamma \odot \frac{h_{i}-\mu_{i}}{\sqrt{\delta_{i}^2 + \varepsilon}}+\beta\right) \\
& \odot\left(1+\operatorname{Linear}_{\gamma}\left(\mathbf{e}^z\right)\right)+\operatorname{Linear}_{\beta}\left(\mathbf{e}^z\right)
\end{aligned}	
\end{equation}
In this equation, $h_{i}$ represents the representation of the $i$-th step within $\mathbf{h}$, 
serving as the input to each Static Fusion Layer. 
Here, $\mu_{i}$ and $\delta_{i}$ are the mean and standard deviation of $h_i$ along the last dimension, 
while $\gamma$ and $\beta$ are the gain and bias parameters that are shared across the sequence, 
enhancing the representational power of the neural network. 
$\varepsilon$ is a small coefficient added for numerical stability.
It is important to note that while the original global factors from layer normalization are retained to maintain the general language distribution, the zero-initialization of $\mathbf{e}^z$ can help to ensure the stability of sequence information during fine-tuning.

\subsection{Data Statistics}
\label{sec:a4}

The statistical information of the HumorWB dataset is illustrated in Tables \ref{tb:a41}, \ref{tb:a42}, and \ref{tb:a43}.
\begin{table}[h]
\centering
\begin{tabular}{lccc}
\toprule[1.5pt]
       & \textbf{All}  & \textbf{Humorless} & \textbf{Humorous} \\
\midrule
All    & 5291 & 3281      & 2010     \\
Train  & 3703 & 2287      & 1416     \\
Valid  & 794  & 499       & 295      \\
Test   & 794  & 495       & 299      \\
\bottomrule[1.5pt]
\end{tabular}
\caption{Number of Posts}
\label{tb:a41}
\end{table}

\begin{table}[h]
\centering
\begin{tabular}{lccc}
\toprule[1.5pt]
       & \textbf{All}   & \textbf{Humorless} & \textbf{Humorous} \\
\midrule[1pt]
All    & 42.34 & 41.48     & 43.75    \\
Train  & 42.56 & 41.65     & 44.04    \\
Valid  & 41.99 & 41.31     & 43.12    \\
Test   & 41.68 & 40.86     & 43.03    \\
\bottomrule[1.5pt]
\end{tabular}
\caption{Average Text Length}
\label{tb:a42}
\end{table}

\begin{table}[h]
\centering
\begin{tabular}{lc}
\toprule[1.5pt]
\textbf{All}    & \textbf{2151} \\
\midrule[1pt]
Train  & 1504 \\
Valid  & 322  \\
Test   & 325  \\
\bottomrule[1.5pt]
\end{tabular}
\caption{Number of Users}
\label{tb:a43}
\end{table}

\subsection{Discussion on Potential Biases in the HumorWB Dataset}
\label{sec:a5}

The HumorWB dataset, which serves as the primary resource for evaluating our proposed CIHR model, is sourced from Weibo, a Chinese social media platform. While the dataset provides a rich combination of humorous texts and detailed user profiles, its origin introduces potential cultural and demographic biases that may limit the generalizability of our findings.

\subsubsection{Cultural Biases}
Humor in the dataset reflects Chinese language and culture, including idiomatic expressions and cultural allusions. While effective for modeling Chinese humor, this cultural specificity may reduce performance on datasets from other linguistic or cultural contexts.

\subsubsection{Demographic Biases} 
Weibo’s user base skews toward younger, urban populations, potentially overrepresenting humor styles popular among these groups while underrepresenting those of older or rural populations.

\subsubsection{Labeling Biases} 
Humor annotations reflect the subjective perceptions of annotators, influenced by their cultural and demographic backgrounds. This may lead to biases in labeling unconventional or less mainstream humor styles.

\end{CJK}
\end{document}